\documentclass{article}
\usepackage{fullpage}

\usepackage[utf8]{inputenc} 
\usepackage[T1]{fontenc}    
\usepackage{url}            
\usepackage{booktabs}       
\usepackage{amsfonts}       
\usepackage{nicefrac}       
\usepackage{microtype}      
\usepackage{soul}
\usepackage{graphicx}
\usepackage{amsmath}
\usepackage{outlines}
\usepackage{xurl}

\usepackage[numbers]{natbib}

\usepackage[colorlinks=true,
citecolor=blue,
filecolor=black,
linkcolor=blue,
urlcolor=blue]{hyperref}

\usepackage{multirow}
\usepackage{paralist}

\usepackage{multicol}
\usepackage{diagbox}

\usepackage[ruled,noend]{algorithm2e}

\usepackage{here}

\usepackage{amsmath,amssymb,amsfonts,amsbsy,amsfonts,latexsym}
\usepackage{makecell}
\usepackage{xcolor}
\usepackage{colortbl}

\usepackage{tabularx,colortbl,xcolor}
\usepackage[normalem]{ulem}
\useunder{\uline}{\ul}{}

\usepackage{wrapfig}

\usepackage{xparse}

\SetKwInput{KwInput}{Input}
\SetKwInput{KwRequire}{Require}

\usepackage{longtable}

\usepackage{chngcntr}
\usepackage{adjustbox}

\begin{document}

\title{Probing Human Visual Robustness with Neurally-Guided Deep Neural Networks}

\author{
Zhenan Shao$^1$ (zhenans2@illinois.edu), Linjian Ma$^1$, Yiqing Zhou$^2$, Yibo Jacky Zhang$^3$, \\Sanmi Koyejo$^3$, Bo Li$^1$, Diane M., Beck$^1$\\
$^1$University of Illinois at Urbana-Champaign, 
$^2$Cornell University, \\
$^3$Stanford University
}

\maketitle


\begin{abstract}
Humans effortlessly navigate the dynamic visual world, yet deep neural networks (DNNs), despite excelling at many visual tasks, are surprisingly vulnerable to minor image perturbations. Past theories suggest that human visual robustness arises from a representational space that evolves along the ventral visual stream (VVS) of the brain to increasingly tolerate object transformations. To test whether robustness is supported by such progression as opposed to being confined exclusively to specialized higher-order regions, we trained DNNs to align their representations with human neural responses from consecutive VVS regions while performing visual tasks. We demonstrate a hierarchical improvement in DNN robustness: alignment to higher-order VVS regions leads to greater improvement. To investigate the mechanism behind such robustness gains, we test a prominent hypothesis that attributes human robustness to the unique geometry of neural category manifolds in the VVS. We first reveal that more desirable manifold properties, specifically, smaller extent and better linear separability, indeed emerge across the human VVS. These properties can be inherited by neurally aligned DNNs and predict their subsequent robustness gains. Furthermore, we show that supervision from neural manifolds alone, via manifold guidance, is sufficient to qualitatively reproduce the hierarchical robustness improvements. Together, these results highlight the critical role of the evolving representational space across VVS in achieving robust visual inference, in part through the formation of more linearly separable category manifolds, which may in turn be leveraged to develop more robust AI systems. 
\end{abstract}
\section{Introduction}
\label{sec:intro}

Deep neural networks (DNNs), despite achieving human-level performance on many visual tasks \citep{deng2009imagenet, xu2015show}, remain brittle under minor image perturbations that are imperceptible and innocuous to humans, such as adversarial attacks \citep{papernot2018sok, szegedy2013intriguing, goodfellow2014explaining}. In contrast, humans are known for their remarkable proficiency in navigating the challenging visual world. A long line of evidence has shown that humans reliably recognize objects across various changes such as translation \citep{biederman1991evidence}, scale \citep{biederman1992size}, viewpoint \citep{biederman1993recognizing}, and even under evolutionarily unnatural image degradations \citep{geirhos2018generalisation} and novel objects that fall outside of typical visual experience \citep{bowers2016visual, vuilleumier2002multiple}. This stark contrast is thought to stem from humans’ ability to develop a unique representational space that supports robust visual inference. DNNs, on the other hand, could potentially lack either the necessary architectural components (such as the extensive feedback and recurrent connections \citep{felleman1991distributed}) or the rich training environment (such as the active interaction with real-world \citep{snow2021treachery}) to naturally achieve such a superior representational space. 

Computational models \citep{serre2007feedforward} and theoretical frameworks \citep{dicarlo2007untangling, iordan2015basic, kravitz2013ventral} have highlighted the ventral visual stream (VVS), comprising hierarchically organized visual cortices such as V1, V2, V4, and the inferotemporal (IT) cortex, as critical for object recognition. Neuroscience studies have shown that representations in higher-order VVS areas become progressively more abstract and invariant to transformations \citep{rust2010selectivity, zoccolan2007trade, isik2014dynamics, hong2016explicit, bao2020map, quiroga2005invariant}. However, these previous studies have typically focused on single regions or limited comparisons between two areas (e.g. V4 vs. IT) \citep{rust2010selectivity, hong2016explicit}, leaving unanswered the fundamental question of \textbf{whether} robust visual representations are indeed progressively built along the VVS hierarchy, and if so, \textbf{how} this progression occurs. An alternative hypothesis is that robustness emerges exclusively in specialized regions such as IT (analogous to how face processing is localized to fusiform gyrus \citep{kanwisher1997fusiform}), without contributions from early or intermediate regions. Resolving this question is important for both providing critical mechanistic insights into how human vision achieves resilience, and informing AI research by revealing the plausibility of intermediate representational stages scaffolding robustness in artificial systems.

To systematically evaluate whether robustness emerges progressively across the VVS, instead of solely in a specialized endpoint like IT, we build on prior work using neural alignment to train DNNs with primate brain data \citep{li2019learning, safarani2021towards, dapello2022aligning, sucholutsky2023getting}. Specifically, we apply ``neural guidance” to steer the final-layer representations of DNNs to match human neural responses, recorded by functional Magnetic Resonance Imaging (fMRI), from consecutive regions of human VVS. By training a separate model aligned to each region, we are able to systematically test the hypothesis that later brain areas confer greater robustness to adversarial perturbations, thus supporting a VVS hierarchy of neural representations with increasing robustness-supporting capacity.

Moreover, by inspecting the resulting more human-like representations, we can also probe the \textbf{mechanism}, and ask what representational properties may be responsible for the improved robustness. One prominent candidate explanation is the manifold disentanglement hypothesis \citep{dicarlo2007untangling, chung2018classification, cohen2020separability}. In this framework, identity-preserving transformations of an object form continuous manifolds in neural representational space, and robustness emerges as these manifolds become less diffuse, lower-dimensional, and more linearly separable across the VVS hierarchy. While theoretically appealing, direct evidence linking human neural manifolds to robustness has been limited. Here, we first test this connection using mean-field theoretic manifold analysis (MFTMA) \citep{chung2018classification}, a recently developed framework that uses statistical mechanical modeling to probe the geometric and statistical properties of manifolds. We measure the extent (i.e., spatial spread) and linear separability of human neural category manifolds across VVS and ask whether such properties can be inherited by DNNs to predict model robustness. We then introduce ``manifold guidance'', a form of category manifold-level neural alignment that moves beyond point-to-point neural guidance to test whether the structure of neural manifolds is indeed one critical component underlying robustness.

Our contributions are as follows:
\begin{itemize}
    \item We show that DNNs trained with neural guidance from later VVS regions show progressively greater robustness to adversarial attacks, corroborating a functional hierarchy across human VVS to support robustness.
    \item We demonstrate that human neural manifolds become less diffuse and more linearly separable across the VVS, and that these properties are inherited by neurally-guided DNNs and predict their subsequent robustness gains.
    \item We introduce ``manifold guidance'' to show that supervision from coarse neural manifold structure is sufficient to qualitatively reproduce the hierarchical robustness effect, supporting neural manifold geometry as a key factor in human visual robustness.
\end{itemize}

\section{Related Work}
\label{sec:related_work}

Neural alignment has been increasingly explored as a method to transfer biological properties to DNNs \citep{sucholutsky2023getting}. Several studies have aligned models to neural responses in order to produce more human-like behavior, such as decision patterns for object recognition tasks \citep{dapello2022aligning, lu2025teaching}, shape-over-texture preferences \citep{khosla2022characterizing}, and even more human-like semantic biases in language models \citep{moussa2025improving}. Work more directly related to robustness used second-order similarity in representational geometry \citep{kriegeskorte2008representational} to regularize DNNs \citep{li2019learning}, targeted early DNN layers \citep{safarani2021towards} or leveraged specialized architectures such as CORnet with layers corresponding to areas like IT \citep{dapello2022aligning}, to demonstrate improved robustness.

However, these studies primarily aim to demonstrate that neural alignment can induce human-like downstream behavior, and thus typically focus on a single brain area, such as V1 or IT, and rely on animal electrophysiology. In contrast, we leverage human fMRI data that allow simultaneous investigation of multiple fine-grained brain regions along ventral stream. This enables us to systematically test whether a gradient of robustness improvement emerges when aligning to progressively higher-order VVS regions. Beyond demonstrating this effect, we further investigate its underlying mechanism, i.e., whether it arises from the geometry of neural category manifolds.

Regarding biological manifolds, Froudarakis et al. \citep{froudarakis2020object} observed that higher-order visual areas in rodents show less diffuse manifolds than earlier ones. Here, we extend this work to demonstrate the decreasing extent and increasing linear separability in human VVS using MFTMA. Prior work \citep{dapello2021neural} has also used MFTMA to show that robust DNNs tend to show less diffuse manifolds, but did not assess whether neural manifold structure itself can induce robustness. We extend these efforts by introducing manifold guidance to directly test whether neural manifold geometry alone is sufficient to confer robustness. Additionally, while MFTMA has inspired unique self-supervised learning methods \citep{yerxa2023learning, schaeffer2024towards}, we use it here as an empirical tool to evaluate whether the robustness-supporting capacity of the human brain can indeed be explained by this theoretical framework.

\section{Methods}
\label{sec:methodology}

\label{methods}

\subsection{Neural guidance}
\begin{wrapfigure}{r}{0.5\textwidth}
\vspace{-1mm}
\centering
\includegraphics[width=.5\textwidth, keepaspectratio]{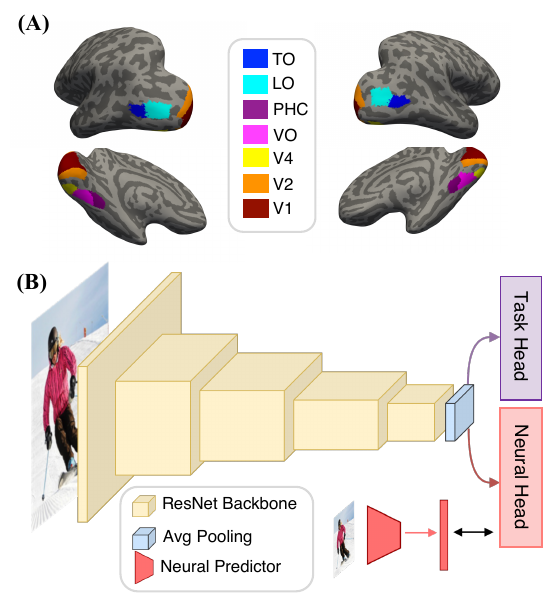}
	\vspace{-6mm}
  \caption{\textbf{(A)} Illustration of anatomical locations of seven ROIs from Subject-1 used for neural guidance. \textbf{(B)} Architecture of the neurally guided ResNet, trained with a task head for image classification and a neural head to match neural responses generated by neural predictors.
  }
  \label{fig:NGmethod}
	\vspace{-1mm}
\end{wrapfigure}

\paragraph{Neural data} 

We used neural data from the publicly available Natural Scenes Dataset (NSD) \citep{allen2022massive}, which provides high-resolution, whole-brain fMRI scans collected from human participants viewing natural images from MSCOCO \citep{lin2014microsoft}. In the NSD experiment, each trial involved participants viewing a single image, and voxel-wise activation levels were estimated via a general linear model (GLM) incorporating a hemodynamic response function regressor and various nuisance regressors for denoising. Activation levels correspond to the GLM coefficients for each image and reflect the percent change in fMRI signal magnitude relative to a voxel’s baseline activity. For each brain region, or Region of Interest (ROI), these values across voxels form a neural representation vector for each image. We used the fully preprocessed data and signal change estimates provided by the NSD processing pipeline to ensure reproducibility. To trace evolving representational spaces across the VVS, we extracted neural responses from seven visual ROIs (Fig.~\ref{fig:NGmethod}A), ranging from early visual areas (V1, V2) to intermediate region (V4), higher-order areas (VO, PHC, LO, TO (see Appendix~\ref{app:roi} for detailed descriptions and expected contributions to robustness). All experiments reported here are based on Subject-1 because optimal registration across individuals remains a significant challenge \citep{haxby2020hyperalignment}. However, using data from other NSD subjects provides similar results, as can be seen in Appendix \ref{app:replicate-sub}.

\paragraph{Neural predictors} 
Following prior neural alignment work \citep{safarani2021towards, dapello2022aligning}, we used the human neural data acquired in NSD to train \footnote{All models in our study were trained using 4 NVIDIA A40 GPUs.} neural predictors, serving as surrogates for each ROI. These neural predictors enabled us to obtain neural responses for any arbitrary images, such as those from the ImageNet dataset, that are not seen by human participants but required for the neural guidance training and subsequent evaluation on DNN robustness. Following the NSD experiments' protocol for train/test split, we divided the 9,841 unique images viewed by Subject-1 into a training set with 8,859 image-representation pairs and a test set with 982 pairs. Separate predictors were trained for each ROI (seven in total) based on a modified ResNet18 architecture \citep{he2016deep}, in which the final fully connected (FC) layer was replaced with a linear layer matching the targeted ROI's  dimensionality (e.g., the V1 predictor output matches the dimensionality of V1 representations). All classification-specific components (e.g., softmax) were removed. Each predictor was trained to minimize a Mean Squared Error (MSE) loss between the predicted and ground-truth neural representations. After training, predictor weights were frozen and used as feature extractors for subsequent neural guidance training. We performed quality control analyses to confirm that predictors indeed captured meaningful neural representations (Appendix~\ref{app:np-qc}).

\paragraph{Neural guidance (NG)} 
As illustrated in Fig.~\ref{fig:NGmethod}B, we adapted ResNet-18 by adding an additional FC layer situated after the last convolutional block. This additional FC layer, or “neural head”, outputs representations matching the dimensionality of the targeted ROI, while the original “task head” performs classification. Training thus jointly optimized the classification and neural representation alignment objectives. Specifically, given input image $x_i$, its corresponding class label $y_i$ and neural representation $r_i$, the shared backbone $f^s$ produces features that are processed by the task head $f^t$ and the neural head $f^n$. The overall training loss is:
\begin{equation} \label{eq:NG}
    L = \alpha L_{\text{CE}}(f^t(f^s(x_i)), y_i) + (1-\alpha)L_{\text{MSE}}(f^n(f^s(x_i)), r_i),
\end{equation}
where $L_{\text{CE}}$ denotes the conventional cross-entropy loss for classification task and $L_{\text{MSE}}$ denotes the MSE loss for matching the representations. We tuned $\alpha$ to achieve an optimal tradeoff between task performance and neural representation alignment.

We trained seven NG-models, each guided by one ROI along the VVS hierarchy to systematically test the effect of neural guidance from different stages of the ventral stream. All models were trained using a curated subset of ImageNet classes (ImageNet-50) selected to closely match the object categories used in NSD. Since MSCOCO images shown to NSD participants do not have a direct one-to-one category mapping to ImageNet, we classified all MSCOCO images used in NSD using a ResNet-101 trained on ImageNet. The 50 most frequently predicted ImageNet categories were selected to define the ImageNet-50 subset. This procedure aimed to maximize the generalizability of neural predictors trained on MSCOCO images in NSD to ImageNet images, thus improving the quality of approximated neural responses. In addition to ImageNet-50 classification, we also trained models with the same neural guidance setup on CIFAR-100 classification and MSCOCO image captioning tasks (thus evaluating changes in BLEU scores). Results for these additional datasets are reported in Appendix \ref{app:replicate-dsets}.

We also included four baseline models each allowing us to rule out different alternative hypotheses. The ``None” model, serving as the absolute baseline condition, had only the task head and was merely fine-tuned on the ImageNet-50 subset. The ``Random” model had the same dual-head structure, but the neural head was guided by an untrained neural predictor with only randomly initialized weights to test the impact of adding an uninformative auxiliary head. The ``V1-shuffle” and ``TO-shuffle” were both dual-headed models guided by the frozen V1 and TO neural predictors, respectively. However, before training, we performed a fixed random reassignment between input images and neural representations, effectively ``shuffling” the correspondence so that each image was consistently paired with an incorrect neural target. This setup tested whether improvements could merely arise from the byproduct of neural-characteristic noise.

\paragraph{Robustness evaluation} \label{sec:methods-att}
During robustness evaluation, the neural head was removed from all NG models, and adversarial attacks were applied solely to the task head. The primary attack method we used here was the untargeted $l_\infty$-based Projected Gradient Descent (PGD) attack \citep{madry2017towards}. We tested a range of attack strengths by sweeping across a range of perturbation bounds $\epsilon$ ($\epsilon = 0.001 * (2i + 1)$, where $i \in \{0, \ldots, 9\}$) to more comprehensively probe the models' performance. To ensure that the observed robustness improvements from neural guidance were not a coincidental result of this attack, we additionally included a range of common adversarial attacks : 1. $l_\infty$-based AutoAttack \citep{croce2020reliable}, which aggregates several attack methods including FAB, Square Attack, APGD-CE, and APGD-DLR; 2. $l_\infty$-based Fast Gradient Sign Method (FGSM) \citep{goodfellow2014explaining}; 3. $l_2$-based Fast Gradient Method (FGM) attack \citep{goodfellow2014explaining}; and 4. $l_2$-based DeepFool attack \citep{moosavi2016deepfool}. Robustness gains under all such attacks are depicted in Appendix~\ref{app:replicate-att}.

\subsection{Neural category manifold disentangling hypothesis}
\paragraph{Manifold statistics} 
To test the manifold disentanglement hypothesis \citep{dicarlo2007untangling}, we first applied MFTMA \citep{chung2018classification} to investigate whether object category manifolds with smaller extent and better linear separability indeed emerge in both human VVS ROIs and neurally guided DNNs. A category manifold is defined as the  representations of all identity-preserving transformations of objects within a category. MFTMA estimates the system's capacity, or linear separability, which reflects the maximal number of such manifolds that can still be linearly separated in an $N$-dimensional space. MFTMA estimates this separability from the geometry of each manifold, specifically its radius $r_c$, which reflects the size of the manifold for category $c$, effective dimension $d_c$ for the number of dimensions with significant variance, and thus extent $e_c = \frac{r_c}{\sqrt{d_c}}$ that captures the spatial diffuseness of the manifold. In particular, a more optimally structured representational space should have manifolds with smaller extent $e_c$ to enable better linear separability. 

We first applied this analysis to human neural representations in each of the seven ROIs using NSD. Because most MSCOCO images in NSD contain multiple objects and thus are not ideal samples for capturing object category manifolds, we curated a subset of images containing only a single object category (see Appendix~\ref{app:mftma-img}), by prioritizing those with large bounding boxes. This resulted in 12 object categories, each with 50 images. We also applied MFTMA on neural representations estimated by the neural predictors using the same subset of images to confirm that they preserve the structure of the original neural manifolds.

To assess manifolds in the NG-models, we used the test set from the ImageNet-50 subset, which provided 50 categories and 50 images per category. To further test the manifold hypothesis \citep{dicarlo2007untangling} that identity-preserving transformations should be treated as part of the same category manifold, we enriched each category with adversarially perturbed variants of the test images, thus obtaining 100 samples in total per category, similar to previous work \citep{dapello2021neural}. These perturbed versions of the test set were generated using the $l_\infty$-based PGD attack with $\epsilon=0.009$.

\paragraph{Manifold guidance} To test whether neural category manifold structure is sufficient to confer robustness, we replaced the neural guidance loss in Eq.~\ref{eq:NG} with constraints that match key geometric properties of DNN category manifolds to those of the corresponding neural ones (see Fig.~\ref{fig:MG}A). Importantly, we approximate each manifold as a first-order linear structure, defined by its center, spread (radius), and local subspace. This simplification allows for efficient integration into DNN training and is sufficient, as a proof-of-concept, to assess whether coarse geometric structure alone can induce robustness improvements.

We first characterized each neural category manifold $c$ offline using neural representations of the training set of ImageNet-50 images (i.e., 50 categories, 1300 clean image samples per category). For each category, we estimated 1. the center $m_c$ as the geometric mean of all samples, 2. the radius $r_c$ as the root-mean-squared (RMS) spread of centered representation vectors and 3. the subspace basis $V_c$ as the top singular vectors from singular value decomposition (SVD) retaining enough components to explaining 99\% of variance.

During training, for each batch of images $X_b$ with labels $Y_b$, we minimize the loss:
\begin{equation} \label{eq:MG}
    L_b = \alpha L_{\text{CE}}(f^t(f^s(X_b), Y_b) + (1-\alpha)\sum_{c\in\text{set}(Y_b)}L_{\text{manifold}}(f^n(f^s(X_{b,c})), m_c, r_c, V_c),
\end{equation}
where $\text{set}(Y_b)$ denotes the unique object categories present in batch $b$, which, given a sufficiently large batch size, typically includes all 50 categories in the dataset, and $L_{\text{manifold}}$ enforces alignment between DNN category manifolds and neural manifolds.
In a given batch, we grouped images based on their category $X_{b,c}$ and treat them as samples for the DNN manifold of category $c$. The manifold guidance loss were computed separately for each category and consisted of two terms: 
\begin{enumerate} 
    \item a radius constraint that penalizes if the DNN manifold radius exceeds the neural one 
\begin{equation}
    L_{\text{RMS}}(Z_{b,c},m_c,r_c) = \max\left(0, f_{\text{RMS}}(Z_{b,c} - m_c) - r_c\right),
\end{equation}
where $Z_{b, c} = f^n(f^s(X_{b,c}))$ and $f_{\text{RMS}}$ denotes the root-mean-squared error,
\item a subspace constraint, analogous to the Variance Accounted For (VAF) \citep{elsayed2016reorganization, gallego2018cortical}, that encourages orientation alignment between the DNN and neural manifolds:
\begin{equation}
    L_{\text{subspace}}(Z_{b,c},V_c) = \frac{ \|Z_{b,c} - V_c V_c^\top Z_{b,c}\|_F^2 }{ \|Z_{b,c}\|_F^2 },
\end{equation}
where $\|\cdot\|_F$ denotes Frobenius norm. 
\end{enumerate}

The total manifold guidance loss is then summed over all categories in the batch
\begin{equation}
    L_{\text{manifold}}(Z_{b,c}, m_c, r_c, V_c) = L_{\text{RMS}}(Z_{b,c},m_c,r_c) + \beta L_{\text{subspace}}(Z_{b,c},V_c),
\end{equation}
where $\beta$ was tuned to balance the weighting between radius and subspace constraints. We note that to reliably estimate DNN category manifolds in each batch, the batch size needs to be large enough to allow a reasonable sample size for each category.

\section{Results}
\label{sec:results}

\subsection{Neural guidance from human VVS induces hierarchical improvement in DNN robustness}

We first tested whether direct neural guidance from later regions of the VVS confers greater adversarial robustness to DNNs. As described above, we trained seven neurally guided DNNs, each using a frozen neural predictor serving as a surrogate for a different ROI along the VVS (V1–TO), while simultaneously learning image classification on ImageNet-50. We additionally included four control models, None, Random, V1-shuffle and TO-shuffle to rule out various hypothesis.

\begin{figure}
  \centering
  \includegraphics[width=1.0\textwidth, keepaspectratio]{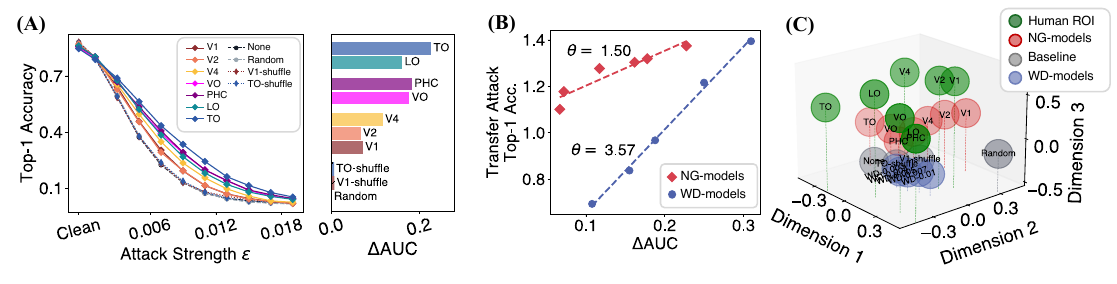}
  \vspace{-5mm}
  \caption{\textbf{(A)} Left: Top-1 classification accuracy of all models under $l_\infty$-based PGD attacks of varying strength $\epsilon$. “Clean” denotes accuracy on unperturbed images. Right: Robustness improvements summarized as the difference in area under the accuracy curve ($\Delta$AUC) for each model relative to the baseline (None model), computed across all $\epsilon$ values used (see Methods~\ref{sec:methods-att}). \textbf{(B)} Transfer attack accuracy vs. native robustness improvement ($\Delta$AUC) for NG- (red diamonds) and WD-models (blue circles). Dashed lines show linear fits with slopes $\theta$ annotated (see Results~\ref{sec:results-surface}). \textbf{(C)} 3D MDS visualization of representational space similarity among 16 models and 7 human ROIs. Proximity indicates similarity in the representational space. Dashed stems are added for visualization of the relative positions of each circle.}
  \label{fig:NGresults}
  \vspace{-2mm}
\end{figure}

Fig.~\ref{fig:NGresults}A shows performance of all models under the $l_\infty$-based PGD attack on the classification head with varying perturbation strengths $\epsilon$. While all baseline models showed similar vulnerability, all NG-models exhibited improved robustness. More importantly, this improvement followed a clear hierarchy. DNNs guided by V1 and V2 showed modest improvement in adversarial robustness, consistent with their early positions in the visual processing hierarchy. Improvements were more pronounced in the V4-guided model and were further surpassed by the LO-guided, VO-guided, and PHC-guided models, with the TO-guided model showing the highest level of robustness. 

We replicated this hierarchical effect across multiple NSD subjects (Appendix~\ref{app:replicate-sub}), alternative tasks and datasets (Appendix~\ref{app:replicate-dsets}), and across multiple attack types (Appendix~\ref{app:replicate-att}). In all cases, we observed a consistent trend: robustness increased with guidance from later visual areas, even though the absolute gains varied across settings. These converging findings across subjects, datasets, tasks, and attacks demonstrate that the robustness conferred by neural guidance is not coincidental. Instead, they suggest that human-constructed visual representations along the VVS become increasingly capable of supporting visual robustness and that such property is transferable to artificial systems.

\subsection{Neural guidance induces unique output surfaces and representational spaces} \label{sec:results-surface}

One possible explanation for the robustness improvement observed in NG-models is that they stem from standard regularization effects. For example, simply increasing weight decay (WD) is a common approach known to improve adversarial robustness by smoothing the model’s output surface, i.e. loss surface with respect to input images \citep{rosca2020case, loshchilov2017decoupled, cohen2019certified, yu2019interpreting}. To assess whether neural guidance offers a distinct solution to robustness, we compared the output surface characteristics of NG-models and DNNs trained with increasing WD regularization.

We first trained five WD models with increasing regularization strength and confirmed that stronger regularization led to higher adversarial robustness (see Appendix~\ref{app:wd}). We also verified that both WD and NG models showed smoother output surfaces via calculating gradient norm magnitude across neighborhood radii \citep{yang2021trs} (see Appendix~\ref{app:smth}). To probe whether the underlying output surface geometry differs between NG and WD models, we leveraged transfer attacks \citep{papernot2016transferability}, which test whether adversarial examples generated for one model can also compromise others. Transferability is a hallmark of conventional DNNs \citep{papernot2016transferability, lu2020enhancing, richards2021adversarial}, suggesting a fundamental homogeneity among them. We first generated adversarial examples from the baseline None model (trained without neural guidance) and evaluated their effectiveness on NG and WD models (see Appendix~\ref{app:transfer-att} for details). Importantly, we matched models based on their native robustness improvement ($\Delta$ AUC, the difference in area under the PGD accuracy curve relative to the None model) to isolate differences in transfer susceptibility. We also fit linear regressions between $\Delta$AUC and transfer attack accuracy separately for each model type. As shown in Fig.~\ref{fig:NGresults}B, NG models were consistently more resistant to transfer attacks than equally robust WD models. We also found that NG models ($\theta = 1.50$) had flatter regression slopes compared to WD models ($\theta = 3.57$), indicating that their transfer susceptibility does not scale as strongly with native robustness. Together, these results suggest that neural guidance alters the output surface in a qualitatively different way compared to common regularization effects.

We hypothesized that these differences in output surfaces arise from changes in representational geometry induced by neural guidance. Using representational similarity analysis (RSA) \citep{kriegeskorte2008representational} (see Appendix~\ref{app:rsa} for methods), we compared internal feature spaces across 16 models (NG, baseline, WD) and 7 human neural ROIs, and derived a Representational Similarity Matrix (RSM) using Spearman’s $\rho$ correlation. We then applied multidimensional scaling (MDS) to project the space similarity structure into 3D space for visualization (Fig.~\ref{fig:NGresults}C). Our results showed that all WD models and baseline models remained highly similar to each other, while NG models diverged from them and shifted closer to human ROIs. This shift in representational geometry suggests that neural guidance steers DNNs toward more human-like internal representations, which may underlie their unique loss landscape structure that ultimately benefited robustness.

\begin{figure}
  \centering
  \includegraphics[width=0.98\textwidth, keepaspectratio]{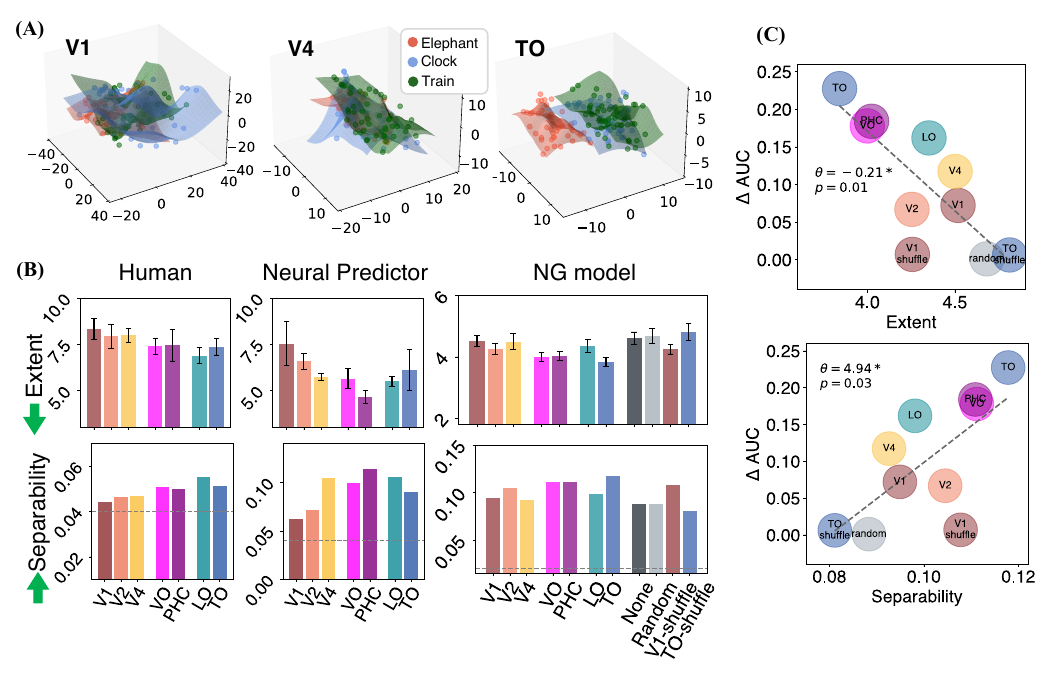}
  \vspace{-3mm}
  \caption{\textbf{(A)} Isomap visualization of three example category manifolds in V1, V4, and TO representational spaces (note the shrinkage in axis scales indicating smaller extent) \textbf{(B)} Category manifold extent ($\downarrow$) and linear separability ($\uparrow$) estimated using MFTMA \citep{chung2018classification} for human brain ROIs (left), neural predictors (middle), and neurally-guided (NG) models (right). Error bars represent 95\% Confidence Interval (CI) of extent across categories. Dashed lines mark the chance level for separability. \textbf{(C)} Manifold extent and linear separability in NG models predict their robustness improvement ($\Delta$AUC): NG models with less diffuse (smaller extent) and more separable manifolds show greater $\Delta$AUC.}
  \label{fig:MFTMA}
  \vspace{-3mm}
\end{figure}

\subsection{Neural manifold geometry predicts robustness in NG models}

Having shown that neural guidance improves robustness and alters representational geometry, we next asked: what specific properties of human neural representations make them particularly effective for supporting robustness? As described above, the manifold disentangling framework \citep{dicarlo2007untangling, chung2018classification} suggests that category manifolds, encompassing all identity-preserving transformations of objects in the given category, are progressively disentangled across the human VVS, naturally leading to robustness. This disentanglement is thought to be achieved by transforming manifolds to have smaller extent and thus become more linearly separable. To illustrate this visually, we show an example in Fig.~\ref{fig:MFTMA}A using Isomap \citep{tenenbaum2000global} to embed three category manifolds into 3D space from V1, V4, and TO representational spaces (see Appendix~\ref{app:mftma-man-vis} for details). As we move up the VVS hierarchy, the manifolds appear progressively less diffuse (note the shrinkage in axis scale) and separated.

Here using MFTMA, we quantified two key manifold statistics: extent, which reflects the diffuseness of each category manifold, and linear separability for all manifolds in the representational space. First, we observed in human neural space that manifold extent decreased and separability increased along the VVS (Fig.~\ref{fig:MFTMA}B left), confirming that later brain regions encode increasingly disentangled category manifolds. Similar patterns of decreasing extent and increasing separability in neural predictors (Fig.~\ref{fig:MFTMA}B middle) confirms that they retain key geometric features of the original neural space.

Similarly, NG models guided by later brain regions showed lower manifold extent and higher separability than those guided by earlier regions (Fig.~\ref{fig:MFTMA}B right). In comparison, baseline models overall showed diffuse, poorly separated manifolds. To quantitatively test whether these manifold properties explain NG models' robustness gains, we fit linear regressions to predict robustness improvement $\Delta$AUC using manifold extent and separability. As expected, models with less diffuse and more linearly separable manifolds showed greater robustness improvements (Fig.~\ref{fig:MFTMA}C). Together, these results show that the desirable manifold properties observed in the human VVS can be inherited by NG models and can predict their downstream robustness gains.

\subsection{Manifold-level guidance qualitatively reproduces robustness hierarchy}
\begin{figure}
  \centering
  \includegraphics[width=0.98\textwidth, keepaspectratio]{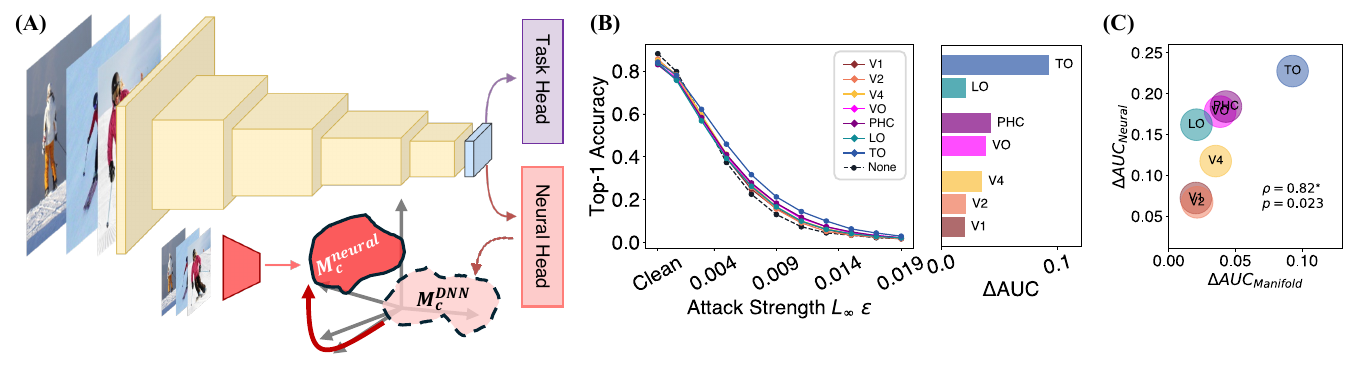}
  \vspace{-1mm}
  \caption{\textbf{(A)} Illustration of manifold guidance that trains DNN with a task head for image classification and a neural head to match category manifold properties with the corresponding neural manifold ones estimated using neural predictors \textbf{(B)} Left: Top-1 classification accuracy of models trained with manifold guidance and the baseline None model under $l_\infty$-based PGD attacks. “Clean” denotes accuracy on unperturbed images. Right: Robustness improvements summarized as $\Delta$AUC relative to the baseline None model. \textbf{(C)} Correlation of robustness improvement $\Delta$AUC for DNNs trained with manifold guidance vs. neural guidance. Improvement hierarchy is preserved despite the small improvement magnitude.}
  \label{fig:MG}
  \vspace{-3mm}
\end{figure}

As proposed by the manifold disentangling hypothesis \citep{dicarlo2007untangling, chung2018classification}, the key factor enabling visual robustness is the linear separability of category manifolds. Therefore, the robustness gains observed in neural guidance may arise from learning the more optimal structure of neural category manifolds, such as their smaller extent, orientation, and resulting separability. Importantly, if the geometry of the manifolds is what matters, then matching only manifold-level properties, instead of precisely matching individual representations, should suffice to recover the hierarchical robustness improvement. Motivated by this, we used a coarser form of supervision, manifold guidance Fig.~\ref{fig:MG}A, to align the overall structure of category manifolds between DNNs and human neural representations. 

As shown in Fig.~\ref{fig:MG}B, models trained with manifold guidance showed a hierarchy in robustness: guidance from later visual areas produced larger gains. Although the overall robustness improvements were smaller than those achieved with full neural guidance, the relative ordering across brain regions was well preserved. Robustness gain ($\Delta$AUC) under manifold guidance was highly correlated with that under neural guidance from the same regions (Spearman's $\rho = 0.82^*, p=0.023$, Fig~\ref{fig:MG}C), indicating that the hierarchical effect from neural guidance is indeed reproduced.

It is important to note that our manifold guidance method uses a linear approximation of manifold geometry and does not capture nuanced nonlinear features such as curvature. Nonetheless, preserving only coarse geometric structure of neural manifolds, such as radius and subspace orientation, without access to any individual neural activation patterns, is already sufficient to recover the hierarchical robustness improvement. This provides strong evidence that learning the neural manifold properties could indeed contribute to the robustness conferred by neural guidance. However, the fact that manifold guidance does not fully reproduce the effects of neural guidance suggests promising directions for future work, such as incorporating more flexible or nonlinear manifold constraints, or investigating the role of finer-grained structures tied to individual neural representations.

\section{Discussion and conclusions}

In our study, we first demonstrated that aligning DNN representations to progressively higher-order regions of the human VVS yields a hierarchical improvement in robustness. We further showed that this progression is not merely a byproduct of generic regularization, but may be driven by the formation of category manifolds with smaller extent and greater linear separability. These results suggest that robustness may emerge through a progressive refinement of representations along the VVS, underscoring the importance of studying the visual system systematically rather than focusing on isolated or endpoint regions. Our findings also point to the geometry of human VVS representations as a potential source of robustness-supporting principles, thus offering promising neural priors and manifold-level constraints to inspire the development of more resilient AI models.

One limitation of our study is that while we demonstrate the relevance of neural manifold structure to the human visual robustness, we do not address how such manifolds are formed in the brain. Although our findings hint at a feed-forward progression across VVS, it is likely that top-down influences, or feedback connections, play a critical role in shaping manifold geometry. The human brain is richly interconnected with  extensive feedback and recurrent connectivity \citep{felleman1991distributed} even for early regions like V1 \citep{lee2003hierarchical}, which may carry higher-level information that extends beyond basic feature detection. This could explain the notable robustness improvements from V1 despite its early position in the visual processing hierarchy. Supporting this idea, recent work has shown the benefits from adding feedback to artificial networks \citep{konkle2023cognitive, shi2023top}. One particularly relevant study demonstrated that feedback in linear neural networks helps reorient manifolds to preserve robustness under noise \citep{naumann2022invariant}. This also aligns with a long-standing view of the visual brain as a generative system that forms and tests perceptual hypotheses informed by both sensory stimulation and environmental statistical regularities \citep{beck2024role, von1925helmholtz}. One intriguing future direction would thus be how specific forms of top-down influence, such as prior knowledge or expectations, guide the development of robust manifold structures and how to incorporating such generative principles in artificial systems.

Interestingly, adversarial attacks on the human visual system have achieved some success, particularly when humans are limited to brief viewing times, reducing the likelihood of engaging feedback connections that most DNN architectures lack \citep{veerabadran2023subtle, gaziv2023strong}. Moreover, monkey IT neurons’ preferences have been shown to be manipulable by adversarial attacks \citep{guo2022adversarially}. Therefore, while our findings affirm that the human visual system likely operates in a representation space with geometries more conducive to robustness, the human visual system may not be entirely impervious to such attacks. In particular, in our experiments, even the best-performing model does not enable DNNs to approach the expected robustness of human performance, as these almost imperceptible adversarial perturbations should not affect human perception at all. This discrepancy could certainly reflect limitations inherent to the quality of fMRI signals, the neural guidance training pipeline, or the choice of ROIs, but it may also indicate that the human visual system itself is not the ultimate gold standard. We leave the investigation into the limits of human visual robustness to future work.

\section*{Acknowledgments}
We thank Dr. Adam Steel for his comments to improve our manuscript. This work used Delta GPU at NCSA  through allocation SOC230011 from the Advanced Cyberinfrastructure Coordination Ecosystem: Services \& Support (ACCESS) program, which is supported by National Science Foundation grants \#2138259, \#2138286, \#2138307, \#2137603, and \#2138296.
{
\bibliographystyle{plain}
\bibliography{ref.bib}
}

\clearpage
\onecolumn
\appendix

\section{Selection of Brain Regions of Interest (ROIs)} \label{app:roi}

We focused on ROIs spanning the VVS from early visual areas to higher-order object-selective cortices along the ventrolateral surface of the brain. Specifically, we selected seven ROIs: V1, V2, V4, VO (Ventral-Occipital), PHC (posterior Parahippocampal Cortex), LO (Lateral-Occipital), and TO (Temporal-Occipital) (illustrated in Fig.~\ref{fig:NGmethod}B). 

V1 and V2 are retinotopically organized areas known for processing low-level details \citep{marr2010vision} and represent the initial stages of the cortical processing hierarchy. V4 serves as an intermediate stage. While not yet exhibiting a pronounced preference for objects, V4 demonstrates early signs of invariance \citep{rust2010selectivity}. Further along the ventral cortex lie VO and PHC. VO is positioned immediately anterior to V4 with a demonstrated object preference \citep{brewer2005visual}, while PHC, located further anterior and overlapping with the Parahippocampal Place Area (PPA) starts to show preference for scenes \citep{arcaro2009retinotopic}, indicative of its capability to extract global information. LO on the lateral surface, which includes LO1 and LO2, overlaps with the relatively posterior section of the functionally defined Lateral Occipital Complex (LOC) \citep{larsson2006two}, commonly recognized for its holistic shape processing capabilities \citep{grill1999differential}. Finally, TO, although less frequently investigated on its own in relation to object recognition, is situated anterior to LO, and potentially encompasses the more anterior portion of LOC \citep{larsson2006two}. 

\begin{wrapfigure}{r}{0.35\textwidth}
\vspace{-2mm}
\centering
\includegraphics[width=.35\textwidth, keepaspectratio]{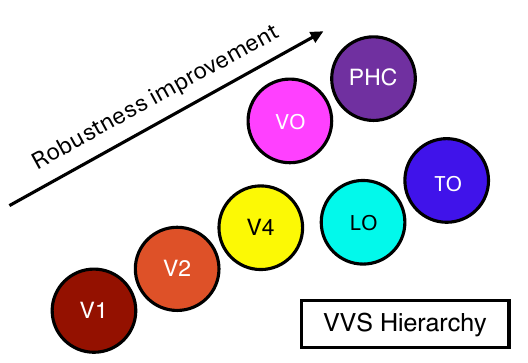}
	\vspace{-5mm}
  \caption{Schematic illustration of expected robustness improvements hierarchy across the seven ROIs included in our study.
  }
  \label{fig:app-hierarchy}
	\vspace{-3mm}
\end{wrapfigure}

Importantly, we do not assume a strictly linear progression of robustness across all higher-order regions. In particular, based on the anatomical positions and functional roles described above, we expect V4 to confer greater robustness than V2 and V1, with all these three early visual areas trailing the higher-level regions. We also expect PHC to confer more robustness than VO due to its scene-level integration, and TO to outperform LO as a more anterior and potentially more abstract object-selective area. However, the relative contributions between the ventral (VO–PHC) and lateral (LO–TO) pairs remain an open question. Therefore, we expect a fork-like hierarchical pattern, where the ventral (VO–PHC) and lateral (LO–TO) ROI pairs may show parallel robustness trends, with a clear progression within each branch (see schematic illustration in Fig.~\ref{fig:app-hierarchy}). This pattern can indeed be observed across our replication experiments (lower rows in Fig.~\ref{fig:app-replicate-sub}, \ref{fig:app-replicate-attacks}, and \ref{fig:app-replicate-dsets}).

\section{Neural predictor quality check} \label{app:np-qc}
We conducted evaluation and control experiments to ensure the quality of the neural predictors used in training NG models. Importantly, we show results from three other NSD subject (Subject-2, 5, and 7) in addition to Subject-1 focused on in the main text, as their data were used for replication experiments in Appendix~\ref{app:replicate-sub}. These four subjects were selected among the eight subjects included in the NSD, because of their completion of the full natural image viewing seccion in NSD, thus providing a more adequate dataset for training the neural predictors. 

\begin{figure}
  \centering
  \includegraphics[width=0.95\textwidth, keepaspectratio]{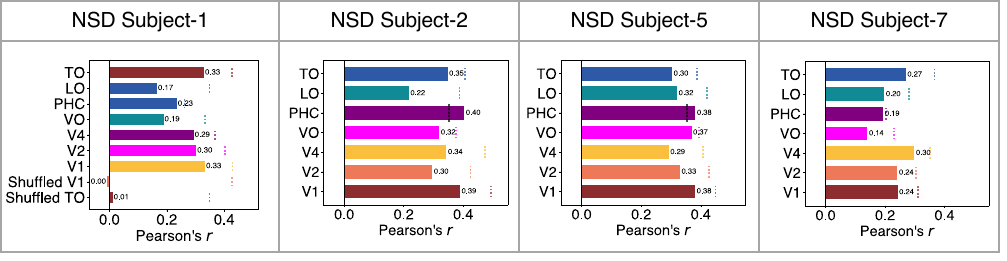}
  \vspace{-2mm}
  \caption{Quality checks of the neural predictors, shown as Pearson’s r correlation between the predicted neural responses and the actual human recordings. Vertical bars indicate the noise-bounded upper limit of neural predictor performance, estimated using split-half correlations with Spearman-Brown correction \citep{yamins2014performance} for each ROI. In Subject-1, we included shuffled conditions (“Shuffled TO” and “Shuffled V1”) served as baseline conditions to show that the neural predictos are indeed learning meaningful information (see Appendix~\ref{app:np-qc}).}
  \label{fig:app-np-qc}
\end{figure}

We evaluated prediction quality by computing Pearson’s $r$ between predicted and actual neural response patterns for each ROI (see Fig.\ref{fig:app-np-qc}). We show the noise ceiling, as vertical bars, estimated using split-half correlations with Spearman-Brown correction \citep{yamins2014performance}, leveraging repeated image presentations in NSD. As can be observed, across all four subjects, neural predictors achieved performance approaching this noise-bounded upper limit.

To confirm that these correlations reflect meaningful learning rather than fitting noises in the fMRI data, we trained control predictors on shuffled data from Subject-1. Specifically, for each input image, we assigned a randomly selected, non-corresponding neural response as the ground truth. However, the integrity of the input-output correspondence was maintained during testing. Therefore, if the correlations can still be observed under these shuffling conditions, then this would suggest that the neural predictors are fitting neural-characteristic noises. However, as shown in Fig.~\ref{fig:app-np-qc}, predictors trained on this shuffled data consistently performed near chance across all ROIs, validating that the original predictors learned informative, image-relevant representations.

We emphasize that these shuffled predictors were used solely for model validation and were discarded before our main analyses. The ``V1-shuffle" and ``TO-shuffle" conditions in the NG model experiments instead used predictors trained on valid neural data (see Methods in the main text for details) .

\section{Replicating robustness improvement hierarchy from Neural Guidance} \label{app:replicate}

\subsection{Across additional NSD subjects} \label{app:replicate-sub}
\begin{figure}
  \centering
  \includegraphics[width=0.85\textwidth, keepaspectratio]{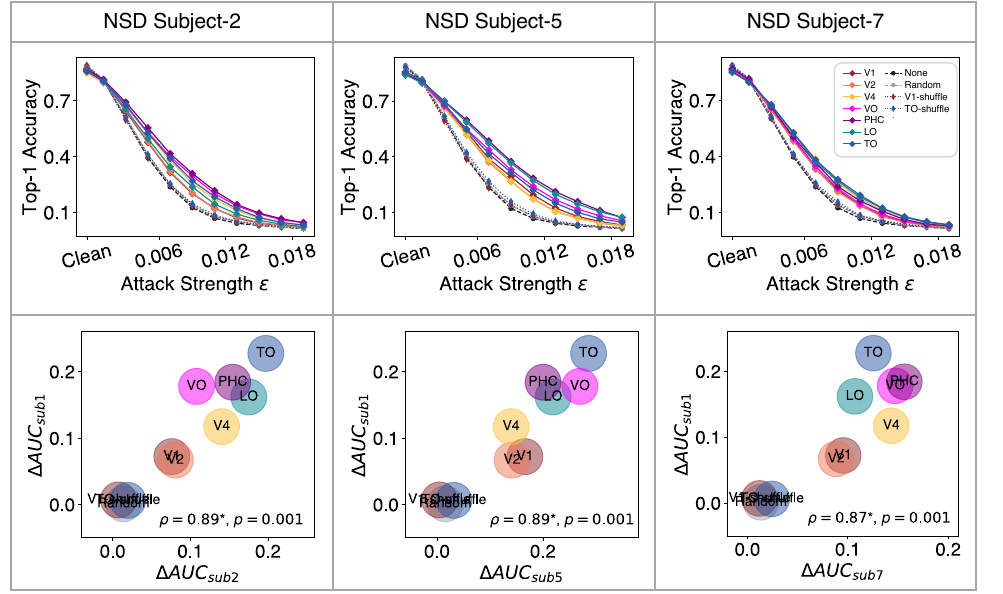}
  \vspace{-2mm}
  \caption{\textbf{Top}: $l_\infty$-PGD attack performance for NG-models guided by Subject-2, 5, and 7 in the NSD dataset. \textbf{Bottom}: The robustness improvement measured in $\Delta$AUC observed for these additional subjects (x-axis) show significant correlation with that of Subject-1 (y-axis), validating the results across individuals.}
  \label{fig:app-replicate-sub}
\end{figure}

To assess the consistency of our findings across individuals, we trained NG-models using three additional NSD participants (Subjects 2, 5, and 7), each of whom, as described above, completed the full natural image viewing sessions and thus allowing sufficient data for training. We evaluated robustness improvements under untargeted $l_\infty$-based PGD attacks. As shown in Fig.\ref{fig:app-replicate-sub}, all three subjects exhibited the same hierarchical trend observed in Subject 1 (main text Fig.\ref{fig:NGresults}): models guided by later visual areas consistently demonstrated greater robustness gains than those guided by early regions.

To quantify the consistency in the robustness hierarchy, we computed $\Delta$AUC between each NG model and the baseline (None) model and assessed the correlation in robustness hierarchies across subjects. Spearman’s $\rho$ values showed statistically significant alignment with the hierarchy observed in Subject 1 (see the lower row of Fig.\ref{fig:app-replicate-sub}). We note that the variability in magnitude is expected, potentially because of both individual differences and variances in the quality of neural data as well as neural predictors, as reflected by noise ceilings and predictor performance in Fig.\ref{fig:app-np-qc}.

\subsection{Across adversarial attacks} \label{app:replicate-att}

We further tested whether the hierarchical robustness improvement observed in NG-models generalizes across different types of adversarial attacks. Using NG-models trained on Subject 1, we evaluated their robustness under four additional attack benchmarks: 1. $l_\infty$-based AutoAttack \citep{croce2020reliable}, which aggregates several attack methods including FAB, Square Attack, APGD-CE, and APGD-DLR; 2. $l_\infty$-based Fast Gradient Sign Method (FGSM) \citep{goodfellow2014explaining}; 3. $l_2$-based Fast Gradient Method (FGM) attack \citep{goodfellow2014explaining}; and 4. $l_2$-based DeepFool attack \citep{moosavi2016deepfool}.

\begin{figure}
  \centering
  \includegraphics[width=0.98\textwidth, keepaspectratio]{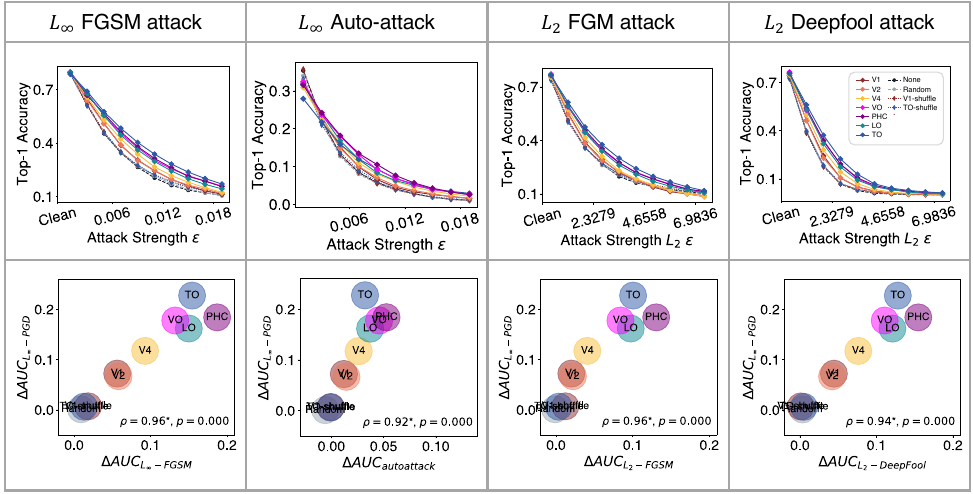}
  \vspace{-2mm}
  \caption{\textbf{Top}: Robustness of NG-models trained with neural data from Subject-1 under four alternative adversarial attacks. \textbf{Bottom}: Robustness improvement hierarchies under these attacks (x-axis) compared to the original $l_\infty$-based PGD results (y-axis) reported in the main text. Spearman’s $\rho$ values indicate significant correspondence across attack types.}
  \label{fig:app-replicate-attacks}
\end{figure}

Despite variations in absolute accuracy, all attack types reproduced the same hierarchical trend: models guided by higher-level ROIs consistently exhibited greater robustness. We quantified this alignment in hierarchy by computing Spearman’s $\rho$ between robustness gains ($\Delta$AUC) from each ROI under each attack and those under the $l_\infty$-based PGD attack reported in the main text. All correlations were statistically significant (Fig.\ref{fig:app-replicate-attacks}, lower row), confirming that the observed hierarchy is robust across a range of attack benchmarks.

\subsection{Across alternate tasks and datasets} \label{app:replicate-dsets}

\begin{figure}
  \centering
  \includegraphics[width=0.65\textwidth, keepaspectratio]{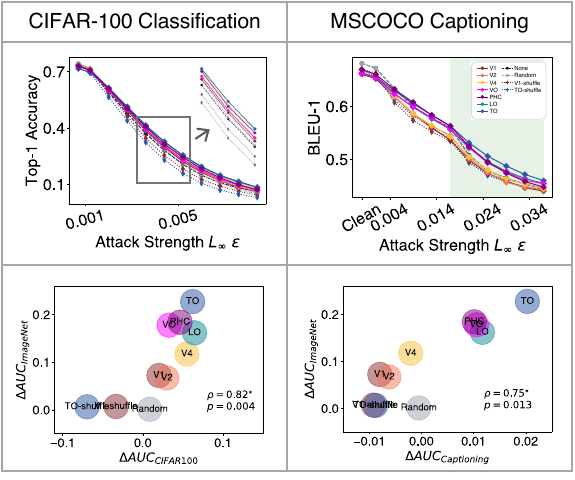}
  \vspace{-2mm}
  \caption{\textbf{Top}: Robustness of NG-models trained with Subject-1 evaluated on CIFAR-100 classification (left) and MSCOCO image captioning (right) tasks under $l_\infty$-based PGD attacks. Black squares and arrows indicate zoomed-in regions for visibility. For the captioning task, the shaded green region highlights the $\epsilon$ range where the hierarchical robustness pattern emerges. \textbf{Bottom}: Robustness improvement hierarchies (measured by $\Delta$AUC) from the two tasks (x-axis) compared to the ImageNet classification results (y-axis) reported in the main text. Significant correlations again confirm the replicability of the hierarchical effect across tasks and datasets.}
  \label{fig:app-replicate-dsets}
\end{figure}

\paragraph{CIFAR-100} 
To assess the effect of neural guidance on a dataset with different visual statistics, we replicated our experiment using another classic classification benchmark, CIFAR-100 \citep{krizhevsky2009learning}. We used Subject-1’s neural data and trained seven additional NG-models, one for each ROI. Given the resolution mismatch with the NSD images (227$\times$227, 32$\times$32 for CIFAR-100), we adapted both the neural predictors and NG-models: we modified the initial convolutional layer (Conv1) of both of the DNNs by reducing both the kernel size (reduced from 7 to 3) and stride (reduced from 2 to 1) to better accommodate the smaller image size. For neural predictors, we trained them with MSCOCO images (those shown to the participants in NSD) that are downsampled to CIFAR-100 resolution for better generalizability.

The NG-models were then evaluated under the same $l_\infty$-based PGD attack. As shown in Fig.~\ref{fig:app-replicate-dsets}, while the overall robustness gains were reduced, potentially due to the limited generalization of neural predictors trained on MSCOCO images to CIFAR-100, the hierarchical pattern of robustness across ROIs was preserved. As in previous experiments, we observed a significant Spearman’s $\rho$ between the $\Delta$AUC values and the VVS hierarchy.

\paragraph{Image captioning with MSCOCO} 
To test whether the robustness benefit of neural guidance generalizes beyond classification, we evaluated the NG-models on an image captioning task. Following standard practice, we used each NG-model’s convolutional backbone (with both heads removed and weights frozen) as a feature extractor, and paired it with a standard LSTM-based decoder equipped with an attention module \citep{xu2015show}. The decoder was initialized with pre-trained weights and then fine-tuned on MSCOCO images viewed by Subject-1 in NSD (the same images used for training the neural predictors). This is to ensure that both the predictors and decoders operate on similar visual content to maximize the transferability. Evaluation was then performed using standard BLEU scores \citep{papineni2002bleu}. To assess robustness, we applied the same $l_\infty$-based PGD attacks used in the classification experiments.

As shown in Fig.~\ref{fig:app-replicate-dsets}, despite overall smaller effect sizes, likely due to the decoder being trained on frozen representations from a classification-focused model, we again observed a consistent robustness hierarchy. NG-models guided by higher-order ROIs produced captions that were more robust to adversarial perturbations, particularly at higher attack strengths (see green region in Fig.~\ref{fig:app-replicate-dsets}), mirroring the patterns observed in classification tasks. Therefore, we were able to show a replication of the robustness improvement hierarchy from neural guidance in the context of image captioning as well.

\section{Uniqueness of NG-models in output surfaces and representational spaces} 

\subsection{WD-models} \label{app:wd}

\begin{wrapfigure}{r}{0.35\textwidth}
\vspace{-2mm}
\centering
\includegraphics[width=.35\textwidth, keepaspectratio]{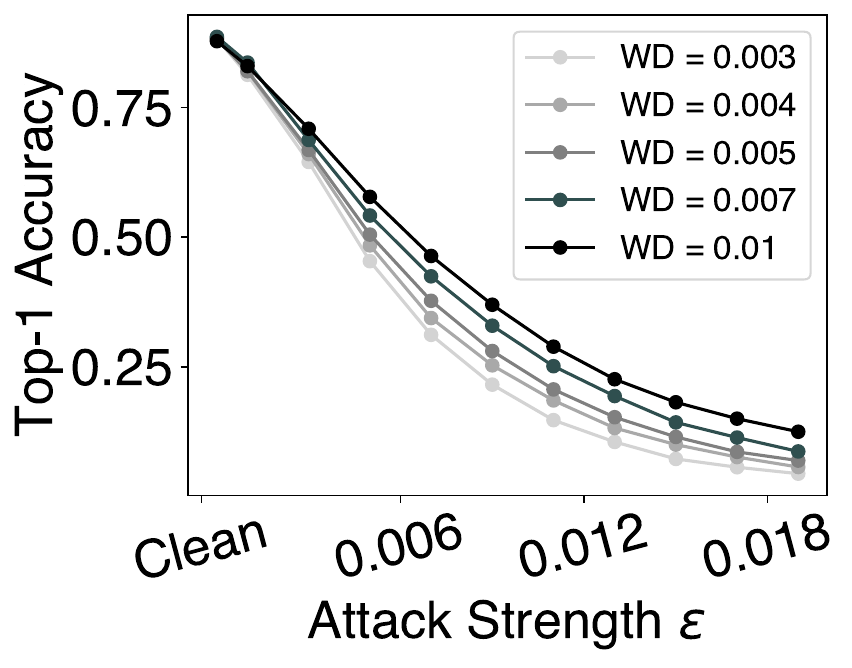}
	\vspace{-5mm}
  \caption{Robustness performance of WD-models trained with five different weight deays values to impose stronger regularization under $l_\infty$-based PGD attacks. The gradient of performance is comparable to those observed with Neural Guidance (Fig.~\ref{fig:NGresults}A).
  }
  \label{fig:app-wd-rob-check}
	\vspace{-3mm}
\end{wrapfigure}

To demonstrate that NG-models achieve a unique output surface and representational space that differs from conventional regularization techniques, we trained additional ResNet-18 models (“WD-models”) on the ImageNet-50 dataset. These models were trained by varying the weight decay parameter \citep{loshchilov2017decoupled}, a common method for imposing regularization \cite{rosca2020case}. For a better comparison, we selected five distinct weight decay values, targeting multiple robustness levels within a relatively comparable range comparable to those observed in neural-guided models, as shown in Fig.~\ref{fig:app-wd-rob-check}.

\subsection{Smoothness check} \label{app:smth}
A key characteristic observed of robust models is their association with smoother output surfaces or more slowly changing loss landscapes with respect to the input images \cite{yu2019interpreting}. Adversarial attacks exploit irregular and complex output surfaces, which facilitates the search for adversarial examples via gradient descent and creates numerous adversarial examples around each correctly classified data sample. Regularization methods such as the straightforward WD manipulation we here adopted \citep{loshchilov2017decoupled, rosca2020case}, along with other more sophisticated defenses \cite{cohen2019certified} all leverage this insight to smooth the neural network function. 

On the other hand, from a representational perspective smooth output surfaces may also arise as a natural consequence of operating on a well-structured representational space. In such a space, semantically similar images—including perturbed instances—are mapped to nearby locations, ideally lying on tighter and disentangled category manifolds \citep{chung2018classification, dicarlo2007untangling}. Linear decision boundaries can then more easily separate category manifolds that include all identity-preserving transformations of object instances. Therefore, we should expect NG-models to also show smoother output surfaces if neural guidance induces more ideally structured representational geometry.

Given an input data $x$, ground truth $y$, and the loss function $\ell_\theta$, we first quantified models' smoothness loss \citep{yang2021trs} as the ``worst" smoothness in a given data point $x$'s $l_\infty$-based neighborhood:

\begin{equation} \label{eq:app-smth}
    L_{\text{smooth}}(\ell, \theta, x, y) = \max_{\|\hat{x}-x\|_\infty} \|\nabla_{\hat{x}}\ell_{\theta} (\hat{x}, y)\|_2
\end{equation}

where $\hat{x}$ is the adversarial image obtained using PGD optimization. We then calculate the gradient norm of the loss with respect to $\hat{x}$. Note that for this quantification, higher values suggest larger gradients, steeper slopes, and therefore less smooth surfaces. For clarity, we applied the following transformation: $e^{-L_{\text{smooth}}}$ so that higher smoothness scores correspond to a smoother output surface.

\begin{figure}
  \centering
  \includegraphics[width=0.85\textwidth, keepaspectratio]{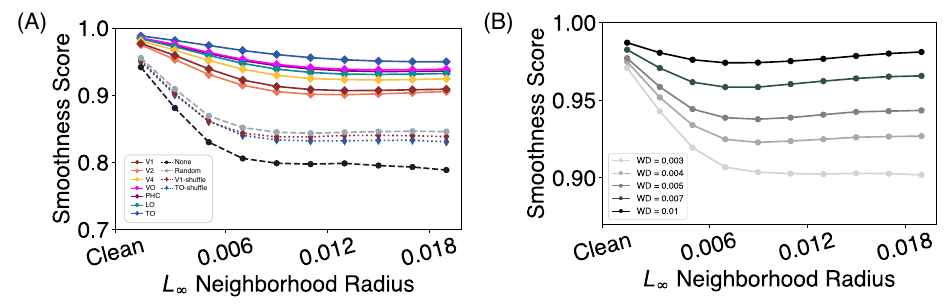}
  \vspace{-4mm}
  \caption{Output surface smoothness scores for (\textbf{A}) NG-models and (\textbf{B})WD-models. Higher values denote smoother output surface.}
  \label{fig:app-smth}
\end{figure}

As shown in Fig.~\ref{fig:app-smth}B, WD-based regularization produced the expected monotonic increase in output surface smoothness with increasing WD magnitude. As predicted, NG models also exhibited enhanced smoothness compared to the baseline None model Fig.~\ref{fig:app-smth}A. Interestingly, even the control models (random, V1-shuffle, TO-shuffle) showed moderate smoothing effects, consistent with the long observed benefits of noise injection \citep{gulcehre2016noisy}. However, all NG-models outperformed these baselines in smoothness across neighborhood radii, with later-region models exhibiting the greatest effects. This hierarchy also largely mirrored that of robustness improvement reported in the main text.

\subsection{Transfer attack} \label{app:transfer-att}

To further assess the similarity of the loss landscape with respect to input images between NG models and the WD models trained with conventional regularization methods, we employed transfer attacks \citep{papernot2016transferability} as an alternative evaluative measure. Specifically, adversarial examples were generated using the $l_\infty$-based PGD attack by attacking the standard “None” model, which had undergone only fine-tuning on the ImageNet-50 dataset. These adversarial examples were then applied to both the set of five WD-models and the seven NG models to evaluate their classification accuracy on these perturbed inputs. \textit{Poor} accuracy then indicates compromised performance and thus a \textit{greater} similarity in the loss landscape, or more precisely, the positioning of adversarial examples, between the source “None” model and the target model. 

\begin{figure}
  \centering
  \includegraphics[width=0.85\textwidth, keepaspectratio]{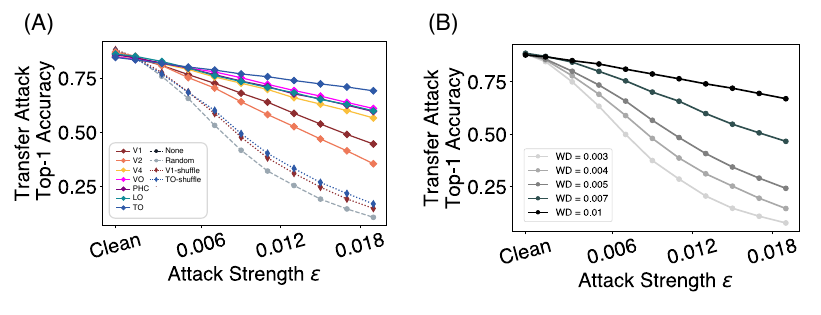}
  \vspace{-4mm}
  \caption{Transfer attack performance (assessed using adversarial examples generated from the None-model—a standard ResNet-18 DNN trained on the ImageNet classification task only) for (\textbf{A}) NG-models and (\textbf{B}) WD-models.}
  \label{fig:app-transfer-att}
\end{figure}

As expected, when attacking NG models using adversarial examples generated from the None model, neurally-guided models showed superior robustness compared to all baseline models (Fig.~\ref{fig:app-transfer-att}A), again with hierarchically later regions exhibiting increasingly better resistance to transfer attacks. Importantly, a similar increase in resistance to the transfer attack was also observed in WD-models (Fig.~\ref{fig:app-transfer-att}B) given that increased smoothness should generally lead to fewer vulnerabilities to be exploited. Such a result might tempt one to consider the hierarchical improvement in robustness observed in NG models as simply a reflection of the increasing regularization. Importantly, however, as we showed in the main text (Fig.~\ref{fig:NGresults}B) with comparable and even lower native robustness, neurally-guided models consistently outperformed WD-models in resisting transfer attacks. Therefore, NG-models developed not only more smooth and thus robust output surface, but also uniquely different ones from conventional regularization effect.

\subsection{Representational Similarity Analysis (RSA)} \label{app:rsa}

We applied RSA \citep{kriegeskorte2008representational} to assess the similarity of representational spaces across all models and human ROIs. To enable direct comparison with human ROIs, we used the MSCOCO images presented during the NSD experiment to construct representational spaces. For all DNNs, we extracted 1D feature vectors from the average-pooling layer for each image. We then computed pairwise Euclidean distances between all image features to form a Representational Dissimilarity Matrix (RDM) for each model. Subsequently, we computed Spearman's $\rho$ correlations between RDMs to produce a Representational Similarity Matrix (RSM), capturing pairwise similarities among the representational spaces of all models and human ROIs. To produce the visualization in Fig,~\ref{fig:NGresults}C, we applied Multidimensional Scaling (MDS) to the RSM, embedding the models into a three-dimensional space where models or ROIs with more similar representational spaces are positioned closer together.

\section{Neural category manifolds} 

\subsection{Selection of images for analyzing human neural manifolds} \label{app:mftma-img}

\begin{figure}
  \centering
  \includegraphics[width=0.98\textwidth, keepaspectratio]{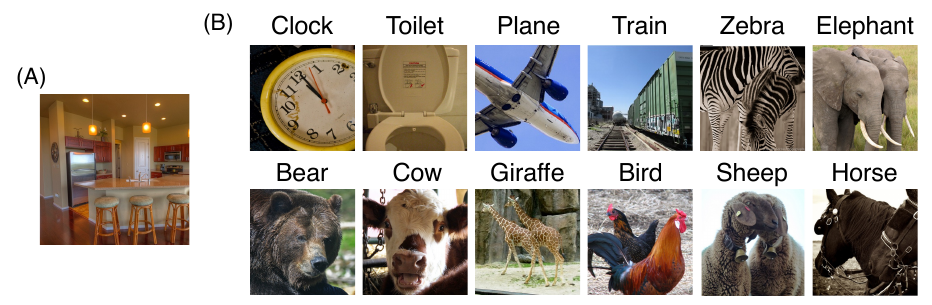}
  \vspace{-2mm}
  \caption{(\textbf{A}): Example MSCOCO image presented to NSD participants containing multiple prominent objects, making it unsuitable for category-level manifold analysis. (\textbf{B}): The 12 object categories selected for human neural manifold analysis using MFTMA, each illustrated with an example image containing a single, dominant object.}
  \label{fig:app-mftma-12categories}
\end{figure}

To estimate category manifold properties, extent and linear separability, using the MFTMA framework \citep{chung2018classification}, we need multiple representative samples per object category. For NG-models trained on ImageNet-50, this process was straightforward: we used the test set with 50 categories and 100 samples per category (50 original and 50 perturbed images). However, for human neural data from the NSD, selecting appropriate category-level samples was more challenging. NSD participants viewed naturalistic images from the MSCOCO dataset \citep{lin2014microsoft}, which is curated for tasks such as object detection and segmentation and often contains complex scenes with multiple overlapping objects (see Fig.~\ref{fig:app-mftma-12categories}A for an example), making them unsuitable for object-centric manifold analysis.

To address this, we curated a subset of images viewed by Subject-1 that were more appropriate for representing single-object categories. Our filtering procedure was as follows: For each image, we identified whether it contained only one object category or whether one object category occupied a significantly larger area (based on aggregated bounding box area) than all other objects combined. If either of the conditions was satisfied, we assigned the image to the dominant object category. Within each selected category, we ranked the images by object size (bounding box area aggregated across all instances of the dominant object category) in descending order and retained the top 50 images per category to serve as category manifold samples. This process yielded a final set of 12 object categories with more appropriate single-object exemplars for MFTMA (see Fig.~\ref{fig:app-mftma-12categories}B for the selected categories and representative examples).

\subsection{Category manifolds visualization} \label{app:mftma-man-vis}

To visually illustrate how category manifolds are transformed across the human VVS, we applied Isomap to neural representations of three example categories (elephant, clock, and train) selected from the 12 filtered categories described above. We performed this separately for V1, V4, and TO, representing early, intermediate, and higher-order regions in the VVS, respectively (see Fig.~\ref{fig:MFTMA}A). 

Following the manifold disentangling hypothesis \citep{dicarlo2007untangling}, which posits that identity-preserving transformations of objects form continuous and separable manifolds, we treated each category as an independent manifold. Isomap was thus applied independently to each category’s high-dimensional neural representations to obtain their 3D embeddings. However, since Isomap embeddings give arbitrary rotation and translation, we registered all manifolds into a common visualization space so that we can compare their relative position, extent and orientation. Specifically, we first extracted a shared 3D basis using the original high-dimensional neural representations with Principal Component Analysis (PCA). We then projected all representations into this common space, and they serve as targeted references for alignment. For each category, we used orthogonal Procrustes analysis to find the optimal rotation that aligned its Isomap embedding to the corresponding projection in the common space. This ensured that manifold shape and internal geometry obtained using Isomap were preserved while aligning all manifolds into the same coordinate system. Finally, each aligned manifold was translated to match the centroid of its corresponding PCA-projected reference and rescaled to match its average spread, thus enabling meaningful visual comparison of scale and spacing.

This visualization illustrates that category manifolds become increasingly compact (smaller extent, as reflected by the decreasing axis ranges in Fig.~\ref{fig:MFTMA}A) and appear more separated from one another in the higher-order visual area TO compared to others. We note that this analysis is intended for illustrative purposes only. For the quantitative assessment of overall manifold properties, we refer readers to the MFTMA-based analyses reported in the main text.

\end{document}